\documentclass[letterpaper, 10 pt, conference]{ieeeconf}  

\IEEEoverridecommandlockouts                              

\title{\LARGE 
Constrained Reinforcement Learning via Dissipative Saddle Flow Dynamics
}
\author{Tianqi Zheng, Pengcheng You, and Enrique Mallada\thanks{T. Zheng and E. Mallada are with the Department of Electrical and Computer Engineering, Johns Hopkins University, Baltimore, MD 21218, USA. Email: {\tt\small \{tzheng8,mallada\}@jhu.edu}. 
P. You is with the Dept. of Industrial Engineering and Management, Peking University, Beijing, China. Email:{\tt\small pcyou@pku.edu.cn}. 
This work was supported by NSF through grants CAREER 1752362, CPS 2136324, and TRIPODS 1934979.
}
}

\usepackage{amsthm}
\usepackage{amsmath}
\usepackage{amsfonts}
\usepackage{amssymb}
\usepackage{algorithm}
\usepackage{algpseudocode}
\usepackage{mathtools}

\usepackage{color,soul}
\allowdisplaybreaks
\usepackage{subfiles}
\usepackage{subfig}
\usepackage{environ}
\usepackage{accents}
\usepackage{balance}
\usepackage{lineno}
\usepackage{graphicx}
\usepackage{subfig}
\usepackage{cite}
 
\usepackage{ifthen}
    \newboolean{arxiv}
    \setboolean{arxiv}{true}

\newtheorem{thm}{Theorem}
\newtheorem{lem}[thm]{Lemma}
\newtheorem{prop}[thm]{Proposition}

\newtheorem{ass}{Assumption}
\allowdisplaybreaks

\let\labelindent\relax 
\usepackage{enumitem}
\setlist[enumerate]{leftmargin=*}
\setlist[itemize]{leftmargin=*}
 
\usepackage{ifthen}
\newboolean{showcomments}
\setboolean{showcomments}{true}
\usepackage[usenames,dvipsnames]{xcolor}
\usepackage{todonotes}

\definecolor{bleudefrance}{rgb}{0.19, 0.55, 0.91}
\definecolor{ao(english)}{rgb}{0.0, 0.5, 0.0}

\newcommand{\addcite}[0]{\ifthenelse{\boolean{showcomments}}
{\textcolor{purple}{(add cite(s)) }}{}}%

\newcommand{\enrique}[1]{  \ifthenelse{\boolean{showcomments}}
{\todo[inline,color=bleudefrance]{Enrique: #1}}{}}
\newcommand{\emmargin}[1]{\ifthenelse{\boolean{showcomments}}{\marginpar{\color{bleudefrance}\tiny EM: #1}}{}}

\newboolean{showedits}
\setboolean{showedits}{true}
\usepackage[xcolor=bleudefrance]{changes}
\definechangesauthor[color=bleudefrance]{EM}
\newcommand{\aem}[1]{
\ifthenelse{\boolean{showedits}}
{\added[id=EM]{#1}}
{#1}
}
\newcommand{\chem}[2]{
\ifthenelse{\boolean{showedits}}
{\replaced[id=EM]{#1}{#2}}
{#1}
}
\newcommand{\dem}[1]{
\ifthenelse{\boolean{showedits}}
{\deleted[id=EM]{#1}}
{}
}

\DeclareSymbolFont{bbold}{U}{bbold}{m}{n}
\DeclareSymbolFontAlphabet{\mathbbold}{bbold}
\DeclareMathOperator*{\argmin}{argmin}

\newcommand{\eucd}{\mathbb{R}}

\usepackage{setspace}
\begin{document}
\maketitle
\begin{abstract}
In constrained reinforcement learning (C-RL), an agent seeks to learn from the environment a policy that maximizes the expected cumulative reward while satisfying minimum requirements in secondary cumulative reward constraints. Several algorithms rooted in sampled-based primal-dual methods have been recently proposed to solve this problem in policy space. However, such methods are based on stochastic gradient descent-ascent algorithms whose trajectories are connected to the optimal policy only after a mixing output stage that depends on the algorithm's history. As a result, there is a mismatch between the behavioral policy and the optimal one. In this work, we propose a novel algorithm for constrained RL that does not suffer from these limitations. Leveraging recent results on regularized saddle-flow dynamics, we develop a novel stochastic gradient descent-ascent algorithm whose trajectories converge to the optimal policy almost surely.

\end{abstract}


\section{Introduction}
Reinforcement learning (RL) studies sequential decision-making problems where the agent aims to maximize its expected total reward by interacting with an unknown environment over time. However, in many applications such as electric grids and robotics, the agent often deals with conflicting requirements \cite{mannor2004geometric}, or has safety constraints during the learning process \cite{achiam2017constrained}. The constrained reinforcement learning (C-RL) framework is a natural way to embed all conflicting requirements efficiently and incorporate safety  \cite{altman1999constrained,paternain2019constrained,castellano2021reinforcement,achiam2017constrained,chen2021primal,bai2022achieving,calvo2021state}. 

There are two major approaches to finding the optimal policy of a C-RL problem, where the first approach solves it in the occupancy measure space. The constrained Markov Decision Process (CMDP) framework is a standard, and well-studied formulation for reinforcement learning with constraints \cite{altman1999constrained}. The agent aims to maximize the total reward function while satisfying requirements in secondary cumulative reward constraints. The CMDP problem can be equivalently written as a linear programming problem in occupancy measure space, and the optimal policy could be recovered from the optimal occupancy measure \cite{altman1999constrained}. However, this approach requires knowledge of the transition kernel of the underlying dynamical system explicitly, which is not always available in many realistic applications. 

An alternative approach is to apply the Lagrangian duality and solve the C-RL problem in policy space \cite{chen2021primal,bai2022achieving,calvo2021state,liu2021learning,ding2020natural}. These approaches solve the min-max optimization problem using a sampling-based primal-dual algorithm or stochastic gradient descent-ascent (SGDA) algorithm, where the Lagrangian function is augmented with a possible regularization term, e.g., a KL divergence regularization. The primal variables and dual variables are updated iteratively, either using gradient information or solving a sub-optimization problem. The outcome of primal-dual algorithms is often subject to two cases: in the first case, the output of the primal-dual algorithm is a mixing policy, which is a weighted average of history outputs \cite{chen2021primal,bai2022achieving,calvo2021state}. In the second case, instead of showing the output policy converges to the optimal policy, they present a regret analysis for objective functions, and constraints \cite{liu2021learning,ding2020natural}. In summary, a key limitation is that the policy often oscillates and does not converge to the optimal policy, i.e., there is a mismatch between the behavioral policy and the optimal one. In this paper, we aim to tackle the above limitations by introducing a novel SGDA algorithm leveraging recent results on regularized saddle flow dynamics. Some of the proofs are omitted due to space constraints.

The key insight that the above sampling-based primal-dual algorithms do not converge is that the Lagrangian function for the C-RL problem does not possess sufficient convexity. The Lagrangian function is bilinear in occupancy measure space and is non-convex-concave in policy space. Our proposed method is rooted in the study of saddle flow dynamics \cite{you2021saddle, cherukuri2016asymptotic}. By adding a carefully crafted augmented regularization, the dissipative saddle flow proposed in \cite{you2021saddle} makes minimal requirements on convexity-concavity and yet still guarantees asymptotic convergence to a saddle point. 

Leveraging tools from this dissipative saddle flow framework, we propose a novel algorithm to solve the C-RL problem in occupancy measure space, where the dynamics asymptotically converge to the optimal occupancy measure and optimal policy. We further extend the continuous-time algorithm in a model-free setting, where the discretized SGDA algorithm is shown to be the stochastic approximation of the continuous-time saddle flow dynamic. We prove that the SGDA algorithm almost surely converges to the optimal solution of the C-RL problem. To the best of our knowledge, this work is the first attempt to solve the C-RL problem to converge to the optimal occupancy measure and policy.

Notation:  Let  $ \mathcal{K} \subset \mathbb{R}^n $ be a closed convex set. Given a point $ y \in \mathbb{R}^n $,  $\Psi_{ \mathcal{K}}[y] = \argmin_{z \in \mathcal{K}} \|z- y\| $ denote the point-wise projection (nearest point) in  $ \mathcal{K}$ to $y$. Given $x \in \mathcal{K}$ and $v \in \mathbb{R}^n$, define the vector field projection of $v$ at $x$ with respect to $  \mathcal{K}$ as:
   $ \Pi_{\mathcal{K}}[x,v] = \lim_{\delta \to 0^+} \frac{\Psi_{ \mathcal{K}}[x + \delta v] -x}{\delta}$

\section{Problem Formulation}
In the constrained reinforcement learning problem (C-RL), $\mathcal{S}$ denotes the finite state space, $ \mathcal{A}$ denotes the finite action space, and $P: \mathcal{S} \times \mathcal{A} \to \triangle^{| \mathcal{S}|}$ gives the transition dynamics of the CMDP, where $P(\cdot | s,a)$ denotes the probability distribution of next state conditioned on the current state $s$ and action $a$. $r: \mathcal{S} \times \mathcal{A} \to [0,1]$ is the reward function, $g^i: \mathcal{S} \times \mathcal{A} \to [-1,1]$ denotes the $i^{th}$ constraint cost function. The scalar $\gamma$ denotes the discount factor, and $q$ denotes the initial distribution of the states.
A stationary policy is a map $\pi: \mathcal{S} \to \triangle^{| \mathcal{A}|}$ from states to a distribution in the action space. The value functions for both reward and constraints' cost following policy $\pi$ are given by:
\begin{align*}
    & V^{\pi}_{r}(q) = (1-\gamma)\mathbf{E}_\pi[\textstyle\sum_{t=0}^\infty \gamma^t r(s_t,a_t) \,|\, s_0\sim q ],\\
    & V^{\pi}_{g^i}(q) = (1-\gamma)\mathbf{E}_\pi[\textstyle\sum_{t=0}^\infty \gamma^t g^i(s_t,a_t) \,|\, s_0\sim q]. 
\end{align*}
The standard C-RL problem aims to maximize the total reward function while satisfying requirements in secondary cumulative reward constraints:
\begin{align}\label{problem:CMDP}
    \max_{\pi}\;\; & V^{\pi}_{r}(q) \nonumber\\ 
    s.t.\;\; & V^{\pi}_{g^i}(q) \geq h^i, \;\;\forall i \in [I] .
\end{align}
There exist two classes of approaches to solving the optimal policy of a constrained reinforcement learning problem. The constrained Markov Decision Process (CMDP) framework equivalently expresses the C-RL problem as a linear programming problem in occupancy measure space \cite{altman1999constrained}. 
Given a policy $\pi$, define $\lambda^{\pi}: \mathcal{S} \times \mathcal{A} \to [0,1]$ as occupancy measure:
\begin{align*}
    \lambda^{\pi}(s,a) = (1-\gamma)\textstyle\sum_{t=0}^\infty \gamma^t P_q^\pi(s_t=s,a_t=a) ,
\end{align*}
where $s_0 \sim q$. By definition, the occupancy measure belongs to the probability simplex $\lambda^{\pi} \in \Delta$. Problem \eqref{problem:CMDP} can be equivalently written as a linear programming problem:
\begin{align}\label{eq:LP_CMDP}
    \max_{\lambda \in \Delta}\;\; &\textstyle \sum_{a} \lambda_a^Tr_a \\
    s.t. \;\; &\textstyle \sum_{a}\lambda_a^T g^i_a \geq h^i, \;\;i \in [I] \nonumber \\
    &\textstyle \sum_{a} (I - \gamma P_a^T)\lambda_a = (1-\gamma)q ,\nonumber    
\end{align} 
where $\lambda_a = [\lambda(1,a),\dots,\lambda(s,a)]^T \in \eucd^{|\mathcal{S} |} $ is the $a^{th}$ column of $\lambda^{\pi}$,  $r_a =[r(1,a),\dots,r(s,a)]^T \in \eucd^{|\mathcal{S} |}$ denotes reward function associated with action $a$, $P_a$ denotes the transition matrix associated with action $a$. The optimal policy could be recovered by finding the optimal occupancy measure 
\ifthenelse{\boolean{arxiv}}{
\begin{align*}
\end{align*}}{$\lambda^* $ from \eqref{eq:LP_CMDP} :
$  \pi^*(a|s) = \lambda^*(s,a)/\sum_{a'\in \mathcal{A}}\lambda^*(s,a')$ \cite{altman1999constrained}. }However, a key limitation in this approach is that it requires knowledge of the transition kernel of the underlying dynamical system explicitly, i.e., $P_a, r_a, g_a^i$.

Another approach is to apply the primal-dual algorithm to find the saddle points of the associated Lagrangian function of problem \eqref{problem:CMDP} in policy space:
\begin{align*}
    L(\pi,\mu) = V^{\pi}_{r} + \textstyle \sum_{i=1}^I \mu_i (V^{\pi}_{g^i} - h^i).
\end{align*}
Algorithms often augment Lagrangian function with a regularization term $ \hat{L}(\pi,\mu) = L(\pi,\mu) + R(\pi,\mu)$, e.g., a KL divergence regularization, and update the policy and dual variable using one of the following rules:
\begin{align*}
    \pi_{k+1}\! =\! \begin{cases}
        \pi_k \! + \! \eta\nabla_{\pi}\hat{L}(\pi,\mu_k) \\ 
        \mathrm{argmax}_{\pi} \hat{L}(\pi,\mu_k)
    \end{cases}
    \mu_{k+1} \!= \!\begin{cases}
        \mu_k \!- \eta\nabla_{\pi}\hat{L}(\pi_k,\mu) \\ 
        \mathrm{argmin}_{\mu} \hat{L}(\pi_k,\mu)
    \end{cases}
\end{align*}

Among the sampling-based primal-dual algorithms, several algorithms output a mixing policy of the form $ \pi_T = \sum_{k=0}^{T-1} \eta_k \pi_k$, which is a weighted average of the history updates \cite{chen2021primal,bai2022achieving,calvo2021state}. The output policy oscillates and does not converge to the optimal policy. On the other hand, several papers provide a regret analysis instead of showing the algorithm's convergence. 
\ifthenelse{\boolean{arxiv}}{
To summarize, the CMDP approach could directly solve the optimal occupancy measure and the optimal policy while requiring knowledge of the transition kernel. The sampling-based primal-dual algorithms often output a mixing policy of history and do not converge to the optimal policy. The key limitation is that the Lagrangian function for the C-RL problem does not possess sufficient convexity. Specifically, the Lagrangian function is bilinear in occupancy measure space and is nonconvex in policy space. In this paper, we aim to provide a novel algorithm that tackles the above difficulties.}
{}

\section{Key insight from saddle flow dynamics}
\begin{figure}[!htb]
    \centering
    \subfloat[\centering time series trajectories]{{\includegraphics[width=4.5cm]{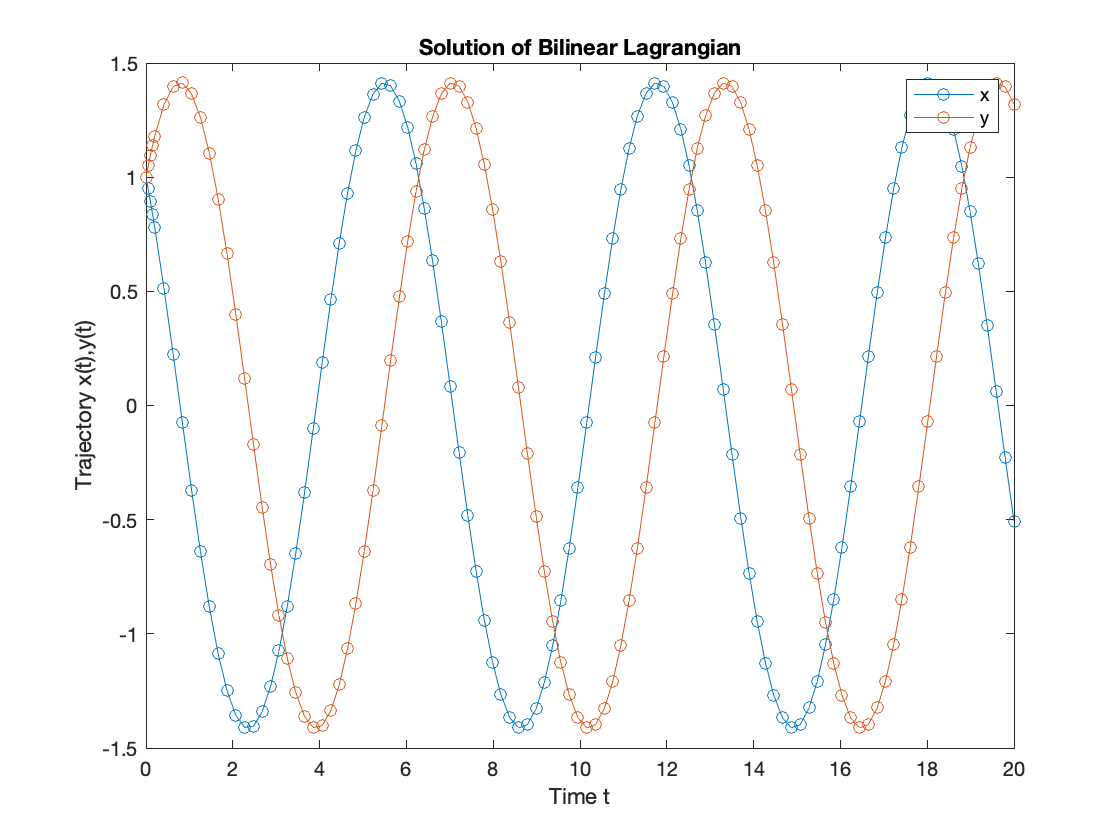} }}%
    \subfloat[\centering phase portrait]{{\includegraphics[width=4.5cm]{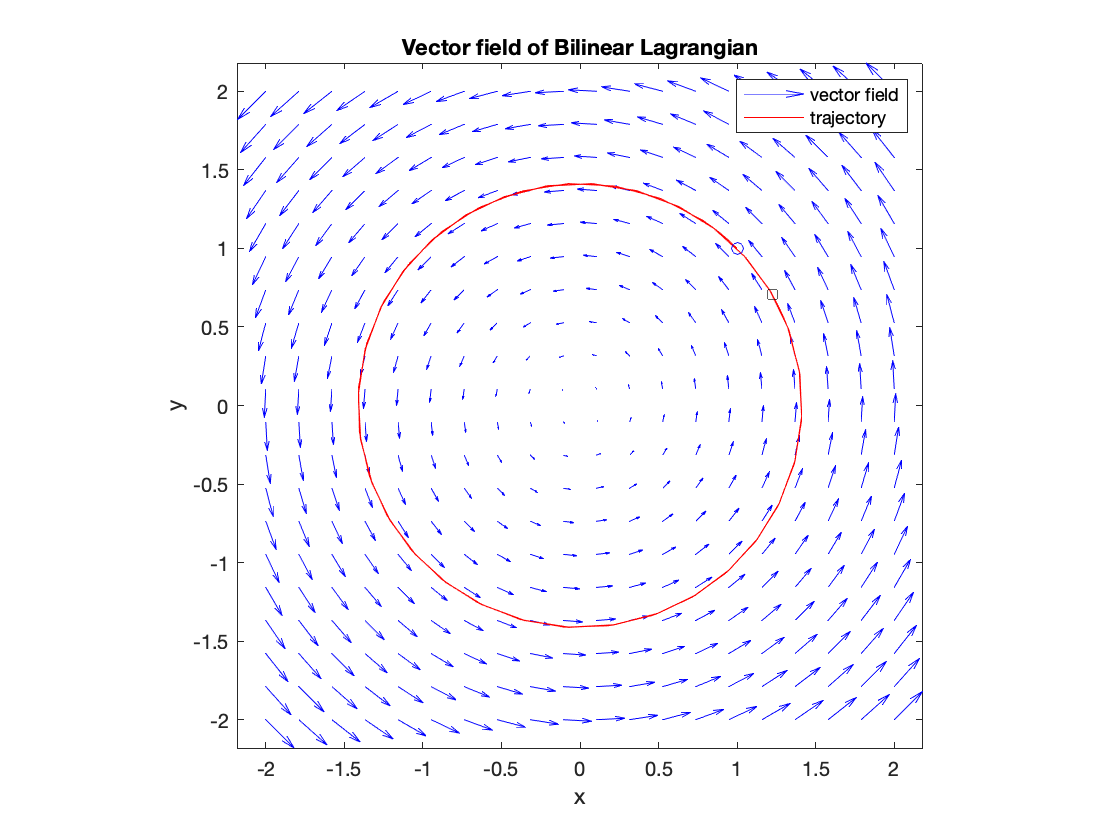} }}%
    \caption{Primal-dual dynamics of bilinear Lagrangian function}%
    \label{fig:Bilinear_example}%
\end{figure}

Before introducing our algorithm, we would like to illustrate our key insight from saddle flow dynamics, which explains why the primal-dual algorithm oscillates and does not converge. For a min-max optimization problem,  primal-dual algorithms require the Lagrangian $L(x,y)$ function to be strictly convex or concave on $x$ or $y$, respectively, to converge. Consider the following motivating example with bilinear Lagrangian function:
\begin{align*}
    \min_{x} \max_{y}L(x,y):=xy .
\end{align*}
Our goal is to apply different dynamic laws that seek to converge to some saddle point $(x^*,y^*)=(0,0)$ of $ L(x,y)$, which satisfies $L(x^*,y) \leq L(x^*,y^*) \leq L(x,y^*) $. In particular, consider the following classical primal-dual algorithm:
\begin{align*}
      & \dot{x} = -\nabla_x L(x,y) = -y,\\
      & \dot{y} = \nabla_y L(x,y) = x.
\end{align*} 
In Figure \ref{fig:Bilinear_example}, (a) plots the time series trajectory of states $x$ and $y$, and (b) plots the vector field and corresponding phase portrait. We observe that the dynamical system oscillates and does not converge to the saddle point (0,0).  

In \cite{you2021saddle}, the authors introduce a regularization framework for saddle flow dynamics that guarantees asymptotic convergence to a saddle point based on mild assumptions. In this paper, we further extend the above framework to solve the C-RL problem.  Specifically, consider the following constrained min-max optimization problem, 
\begin{align*}
\min_{x \in \mathcal{K}} \max_{y \in\mathcal{V}}L(x,y)
\end{align*} where $\mathcal{K} \subset \eucd^n,\mathcal{V} \subset \eucd^m$ are bounded closed convex sets. We propose a regularized surrogate for $L(x,y)$ via the following augmentation:
\begin{align*}
    L(x,y,z,w) :=\frac{1}{2\rho}\|x-z \|^2+ L(x,y)-\frac{1}{2\rho}\|y-w \|^2
\end{align*}
The following projected and regularized saddle flow dynamics aim to find the saddle points of the regularized Lagrangian, which contains the saddle point  of the original Lagrangian. The regularized saddle flow dynamics still preserve the same distribution structure, which can be implemented in a fully distributed fashion, and requires the same gradient information as the classical primal-dual algorithm:
\begin{align}\label{eq: projected saddle flow dynamics subequations}
&\dot{x} = \Pi_{\mathcal{K}}\Bigr[x, \!-\!\nabla_xL(x,y) \!-\! \frac{1}{\rho}(x-z)\Bigr] , \dot{z} =\Pi_{\mathcal{K}}\Bigr[z, \frac{1}{\rho}(x-z)\Bigr] \nonumber \\ 
&\dot{y} = \Pi_{\mathcal{V}}\Bigr[y,\!-\!\nabla_yL(x,y) \!-\! \frac{1}{\rho}(y-w)\Bigr] , \dot{w} =\Pi_{\mathcal{V}}\Bigr[w, \frac{1}{\rho}(y-w) \Bigr]  
\end{align}

\begin{thm}\label{thm:Saddle FLow Dynamics}
Assume that $L(\cdot, y)$ is convex for $\forall y$ and $L(x,\cdot)$ is concave for $\forall x$, continuously differentiable, and there exists at least one saddle point $(x^* \in \mathcal{K} ,y^* \in \mathcal{V})$, where $\mathcal{K} \subset \eucd^n,\mathcal{V} \subset \eucd^m$ are closed and convex. Then the projected saddle flow dynamics \eqref{eq: projected saddle flow dynamics subequations} asymptotically converge to some saddle point $(x^*,y^*)$ of $L(x,y)$, while $x(t) \in \mathcal{K}, y(t) \in \mathcal{V}, \forall t$ with initialization $x(0) \in \mathcal{K}, y(0) \in \mathcal{V} $.

\ifthenelse{\boolean{arxiv}}{
\textit{Proof}: See Appendix}
{\textit{Proof}: See \cite{zheng2022constrained}.}
\end{thm}

The above theorem shows the projected and regularized saddle flow dynamics will asymptotically converge to the saddle point of the Lagrangian function, which requires mild assumptions on convexity. Additionally, the following result summarizes conditions under which the solutions of the projected system exist and are unique.

\begin{prop}\cite[Prop 2.2]{cherukuri2016asymptotic} \label{pro:existence_projected_system}
Let $f: \eucd^n \to \eucd^n$ be Lipschitz on a closed convex polyhedron $\mathcal{K} \in \eucd^n $. Then, for any $x_0 \in \mathcal{K} $, there exists a unique solution $t \to x(t)$ of the projected system $\dot{x} = \Pi_{\mathcal{K}}\Bigr[x, f(x)\Bigr]$ with $x(0) = x_0$.
\end{prop}

\begin{figure}[!htb]
\centerline{\includegraphics[width=0.6\columnwidth]{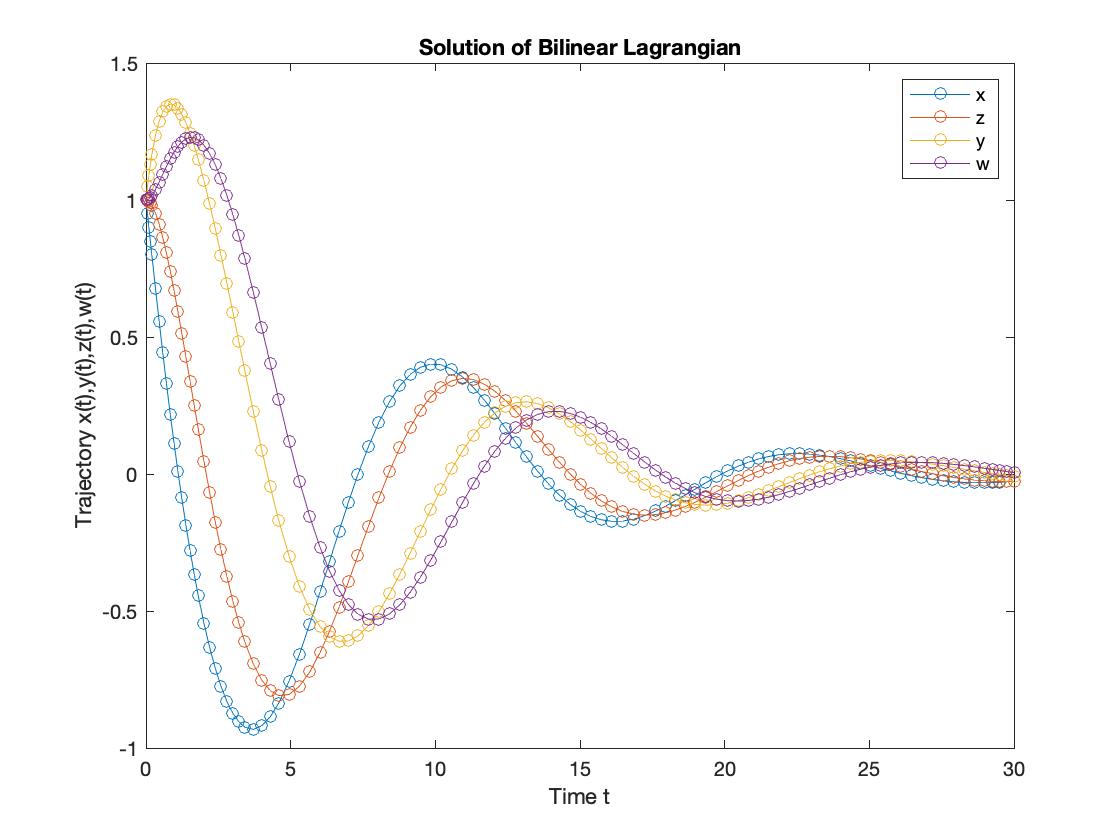}}
   \caption{Regularized saddle flow dynamics for $ L(x,y)=xy$ }
\label{fig:2}
\end{figure}

We now apply the regularized saddle flow dynamics to the bilinear Lagrangian function $L(x,y)=xy $.
\ifthenelse{\boolean{arxiv}}{ 
\begin{subequations}
\begin{align*}
& \dot{x} =-y-\frac{1}{\rho}(x-z), & \dot{z} = \frac{1}{\rho}(x-z), \\
& \dot{y} = x - \frac{1}{\rho}(y-w), & \dot{w} = \frac{1}{\rho}(y-w).
\end{align*}
\end{subequations}
}{}According to Figure \ref{fig:2}, the trajectories of the above saddle flow dynamics asymptotically converge to the saddle point $(0,0,0,0)$, even when the original Lagrangian function is bilinear.

A direct application of the above projected and regularized saddle flow dynamic is to solve the C-RL problem in occupancy measure space \eqref{eq:LP_CMDP}, where the Lagrangian function is also bilinear. Specifically, the Lagrangian function for \eqref{eq:LP_CMDP} in occupancy measure space is:
\begin{align}\label{eq:Lagrangian_LP_CMDP}
    L(\lambda,\mu,v)= &\sum_{a} \lambda_a^Tr_a + \sum_{i}\mu_i(\sum_{a}\lambda_a^T g^i_a - h^i)\nonumber \\ & +(1-\gamma) \langle q,v\rangle - \sum_{a\in \mathcal{A}}\lambda_a^T (I - \gamma P_a)v,
\end{align} where $\mu_i \geq 0$ is the Lagrange multiplier associated with the $i^{th}$ inequality constraint and $v$ is the Lagrange multiplier associated with the equality constraint. Therefore, motivated by the projected and regularized saddle flow dynamics framework, we propose a regularized surrogate for \eqref{eq:Lagrangian_LP_CMDP} via the following augmentation:
\begin{align}\label{eq:augmented lagrangian}
     L(v,\hat{v},\mu,\hat{\mu},\lambda,\hat{\lambda})& := \frac{1}{2\rho}\|v-\hat{v} \|^2+ \frac{1}{2\rho}\|\mu-\hat{\mu} \|^2 \nonumber \\ &+ L(v,\mu,\lambda) -\frac{1}{2\rho}\|\lambda-\hat{\lambda} \|^2 
\end{align}
Slater's condition for C-RL and the following Lemma establishes the boundedness of dual decision variables, which naturally provides a closed convex set for projection. 

\begin{ass}[Slater's condition for C-RL]\label{ass:Slater C-RL}
There exists a strictly feasible occupancy measure $\tilde{\lambda} \in \Delta$ of problem \eqref{eq:LP_CMDP}, i.e., there exist some $\psi > 0$ such that
\begin{align*}
& \sum_{a}\tilde{\lambda}_a^T g^i_a \geq h^i + \psi,\;\; i \in [I] \nonumber \\
& \sum_{a\in \mathcal{A}} (I - \gamma P_a^T)\tilde{\lambda}_a = (1-\gamma)q ,\nonumber       
\end{align*}
\end{ass}

\begin{lem}\cite[Lem.1]{bai2022achieving}[Bounded dual variable]
Under the assumption \ref{ass:Slater C-RL}, the optimal dual variables $\mu^*,v^*$ are bounded. Formally, it holds that $\| \mu^* \|_1 \leq \frac{2}{\psi}$ and $\| v^* \|_{\infty} \leq \frac{1}{1-\gamma}+\frac{2}{(1-\gamma)\psi}$
\end{lem}

Therefore, we propose the following projected saddle flow dynamics to find the saddle points of \eqref{eq:augmented lagrangian}, where $\mathcal{U}:= \{\mu | \mu \in \eucd^{I}_{\geq 0}, \| \mu \|_1 \leq \frac{2}{\psi} \}, \mathcal{V}:= \{ v | v \in \eucd^s, \| v^* \|_{\infty} \leq \frac{1}{1-\gamma}+\frac{2}{(1-\gamma)\psi} \}$ are both closed convex polyhedrons.
\begin{align}\label{eq:regularized saddle flow dynamic C-RL}
&\Dot{v} \;= \Pi_{\mathcal{V}}\Bigr[v, \sum_{a\in \mathcal{A}} (I - \gamma P_a^T)\lambda_a-(1-\gamma)q - \frac{1}{\rho}(v-\hat{v})\Bigr] \nonumber ,\\
&\Dot{\hat{v}} \;= \Pi_{\mathcal{V}}\Bigr[\hat{v},\frac{1}{\rho}(v-\hat{v}) \Bigr]\nonumber,\\ 
&\Dot{\mu}_i = \Pi_{\mathcal{U}}\Bigr[\mu_i, h^i- \sum_{a}\lambda_a^T g^i_a - \frac{1}{\rho}(\mu_i-\hat{\mu}_i)\Bigr] \nonumber,\\
&\Dot{\hat{\mu}}_i = \Pi_{\mathcal{U}}\Bigr[\hat{\mu}, \frac{1}{\rho}(\mu-\hat{\mu})\Bigr] \nonumber,\\
&\Dot{\lambda}_a =\Pi_{\Delta}\Bigr[\lambda, r_a - (I - \gamma P_a)v +  \sum_{i}\mu_i g^i_a- \frac{1}{\rho}(\lambda_a-\hat{\lambda}_a)\Bigr]  \nonumber,\\
&\Dot{\hat{\lambda}}_a =\Pi_{\Delta}\Bigr[\hat{\lambda}_a, \frac{1}{\rho}(\lambda-\hat{\lambda})\Bigr] ,
\end{align}

The following theorem is a direct application of Theorem \ref{thm:Saddle FLow Dynamics} and Proposition \ref{pro:existence_projected_system}, which guarantees \eqref{eq:regularized saddle flow dynamic C-RL} asymptotically converge to the unique (optimal) saddle point of the C-RL problem \eqref{eq:LP_CMDP}. Then we could recover the optimal policy from the optimal occupancy measure $\lambda^*$.

\begin{thm}\label{Thm:Saddle flow for C-RL}
Let Assumption \ref{ass:Slater C-RL} hold. Then the projected saddle flow dynamics \eqref{eq:regularized saddle flow dynamic C-RL} asymptotically converge to some saddle point $(\lambda^* ,\mu^* ,v^*)$ of $L(\lambda,\mu,v)$, while satisfying $\lambda(t)\in \Delta, \mu(t) \in \mathcal{U}, \forall t$ with proper initialization.
\end{thm}

\section{Stochastic Approximation for C-RL}
In the following section, we aim to extend the proposed continuous-time saddle flow algorithm \eqref{eq:regularized saddle flow dynamic C-RL} to a model-free setting. Specifically, we propose a novel stochastic gradient descent-ascent algorithm, which does not require the knowledge of transition kernel. We show that the SGDA algorithm is a stochastic approximation of the continuous time saddle flow dynamics \eqref{eq:regularized saddle flow dynamic C-RL}, which almost surely (w.p.1) converges to the unique saddle point of the C-RL problem.

In many optimization problems, the goal is to find some recursive numerical procedure that sequentially approximates a value of the decision variable $x$, which minimizes the objective function, e.g., $\dot{x} =h(x)$ or $  x^{n+1} = x^n + \alpha^nh(x^n)$. Stochastic approximations attempt to solve the problem when one cannot actually observe $h(x)$, but rather $h(x)$ plus some error or noise. Consider the following projection algorithm:
\begin{align}\label{eq:projection algorithm from projection theorem}
x^{n+1} = \Psi_{ \mathcal{G} }\Bigr[x^n + \alpha^n \Bigr(h(x^n) + \xi^n\Bigr) \Bigr],  
\end{align}
where $\mathcal{G}  := \{x: q_i(x) \leq 0, i \in [s]  \} $ denotes the constraints and $\{\xi^n \}$ denotes a sequence of random variables. The goal is to generate a sequence $\{x^n\}$ estimate of the optimal value of $x$ when the actual observation has random noise $h(x^n) + \xi^n$. In general, the projection $\Psi_{ \mathcal{G} }[x]$ is easy to compute when the constraints are linear; i.e., when $ \mathcal{G} $ is a polyhedron. We introduce the following list of standard assumptions for stochastic approximation
\begin{ass}[Stochastic Approximation]\label{Ass: Stochastic approximation} \;\;
\begin{enumerate}
\item[1.1] $h(\cdot)$ is a continuous function.
\item[1.2] $\{\alpha^n\} $ is a sequence of positive real numbers such that $\alpha^n >0, \sum_n \alpha^n = \infty,  \sum_n(\alpha^n)^2 < \infty$,
\item[1.3] G is the closure of its interior and is bounded. The $q_i(\cdot), i \in [s]$ are continuously differentiable.
\item[1.4] There is a $T>0$ such that for each $\epsilon > 0$
\begin{align*}
    \lim_n P\{\sup_{j \geq n} \max_{t \leq T}|\sum_{i=m(jT)}^{m(jT+t)-1}\alpha^i\xi^i |\geq \epsilon \}=0,
\end{align*}
where $t^n := \sum^{n-1}_{i=0} \alpha^i$ and $ m(t):= \max_n\{ t^n \leq t\}$ for $t \geq 0$.
\end{enumerate}
\end{ass}

Under those standard assumptions for stochastic approximations, the sequence  $\{x^n \}$ generated by the projection algorithm \eqref{eq:projection algorithm from projection theorem} will converge almost surely to a stable solution to the projected system.

\begin{thm}\cite[Theorem 5.3.1]{kushner2012stochastic}\label{Thm: Projection Theorem}
Assume Assumption \ref{Ass: Stochastic approximation} hold. Consider the following ODE:
\begin{align}\label{eq:Projected ODE projection theorem}
    \dot{x} = \Pi_{\mathcal{G}}\Bigr[x,h(x)\Bigr].
\end{align} Let $x^*$ denotes an asymptotically stable point of \eqref{eq:Projected ODE projection theorem} with domain of attraction $DA(x^*)$ and $x^n$ generated by \eqref{eq:projection algorithm from projection theorem}.
If $A \in DA(x^*)$ is compact and $x^n \in A$ infinitely often, then $x^n$ converges to $x^*$ almost surely as $n \to \infty$.
\end{thm}

Consider the following randomized primal-dual approach proposed in \cite{bai2022achieving,wang2020randomized}, where we assume the presence of a generative model. For a given state action pair $(s,a)$, the generative model provides the next state $s' $ and the reward functions $r(s,a), g^i(s,a)$ to train the policy. Consider the following stochastic approximation for the Lagrangian function \eqref{eq:Lagrangian_LP_CMDP} for a distribution $ \xi $:
\begin{align} \label{eq:Stochastic approximation for Lagrangian function}
& L^{\xi}(\lambda,\mu,v)=(1-\gamma)v(s_0) - \sum_{i \in [I] } \mu_ih^i +\\ & \mathbf{1}_{\xi(s,a)>0}\frac{\lambda(s,a)\Bigr[r(s,a)-v(s)+\gamma v(s')+\sum_{i \in [I]}\mu_i g^i(s,a) \Bigr]}{\xi(s,a)}  \nonumber
\end{align}
where $s_0 \sim q, (s,a) \sim \xi $, and the next state $s' \sim P(\cdot | s,a)$. The stochastic approximation $ L^{\xi}(\lambda,\mu,v)$ \eqref{eq:Stochastic approximation for Lagrangian function} is an unbiased estimator for the Lagrangian function \eqref{eq:Lagrangian_LP_CMDP}, i.e., $\mathbf{E}_{\xi, P(\cdot | s,a),q }\Bigr[L^{\xi}(\lambda,\mu,v)\Bigr]  = L(\lambda,\mu,v) $. Using the proposed stochastic approximation of the Lagrangian function, consider the following projection algorithm for solving the C-RL problem in a model-free setting:


\begin{align}\label{eq:SA regularized saddle flow dynamic}
v^{n+1} &= \Psi_{\mathcal{V}}\Bigr[ v^n+\alpha^n\Bigr(\mathbf{1}_{\xi(s,a)>0}\frac{\lambda(s,a)[e(s)-\gamma e(s')]}{\xi(s,a)} \nonumber \\ &-(1-\gamma)\mathbf{e}(s_0)- \frac{1}{\rho}(v^n-\hat{v}^n)\Bigr) \Bigr] , \nonumber \\
\hat{v}^{n+1} &= \Psi_{\mathcal{V}}\Bigr[\hat{v}^{n}+\alpha^n \frac{1}{\rho}(v^n-\hat{v}^n) \Bigr], \nonumber\\ 
\mu_i^{n+1} &= \Psi_{\mathcal{U}}\Bigr[\mu_i^n + \alpha^n\Bigr(h^i -
\mathbf{1}_{\xi(s,a)>0}\frac{\lambda(s,a)g^i(s,a) }{\xi(s,a)} \nonumber \\
& - \frac{1}{\rho}(\mu_i^n-\hat{\mu}_i^n)\Bigr)\Bigr], \nonumber\\
\hat{\mu}_i^{n+1} &= \Psi_{\mathcal{U}}\Bigr[\hat{\mu}_i^{n}+\alpha^n \frac{1}{\rho}(\mu_i^n-\hat{\mu}_i^n)\Bigr], \nonumber\\
\lambda_a^{n+1} &=\Psi_{ \Delta}\Bigr[ \lambda_a^{n}+\alpha^n\Bigr(-\frac{1}{\rho}(\lambda_a^n-\hat{\lambda}_a^n)  \nonumber \\ & + \mathbf{1}_{\xi(s,a)>0}\frac{ r(s,a) - v(s)+\gamma v(s')+ \sum_{i}\mu_i^n g^i(s,a)}{\xi(s,a)}  \Bigr)\Bigr]  , \nonumber\\
\hat{\lambda}_a^{n+1} &= \Psi_{ \Delta} \Bigr[\hat{\lambda}_a^{n} + \frac{1}{\rho}(\lambda_a^n-\hat{\lambda}_a^n) \Bigr],
\end{align}

The following Theorem is a direct application of Theorem \ref{Thm: Projection Theorem} and  \ref{Thm:Saddle flow for C-RL}, which shows the sequence from \eqref{eq:SA regularized saddle flow dynamic} almost surely converges to the optimal solution to the C-RL problem.
\begin{thm}
    Assume \ref{ass:Slater C-RL} and \ref{Ass: Stochastic approximation} hold, as $n \to \infty$, the sequence $\{\lambda^n, v^n ,\mu^n\} $ generated by \eqref{eq:SA regularized saddle flow dynamic} almost surely (w.p.1) converge to the optimal solution of the C-RL problem \eqref{eq:LP_CMDP}.
\end{thm}

\section{Numerical Examples} \label{sec:numerical}
In this section, we illustrate the effectiveness of our proposed approach using a classical CMDP problem: flow and service control problem in a single-server queue \cite{altman1999constrained}. Specifically, we consider a discrete-time single-server queue with a buffer of finite size $L$. We assume that, at most, one customer may join the system in a time slot. The state $s$ corresponds to the number of customers in the queue at the beginning of a time slot $ (|\mathcal{S}| = L+1) $. 
The service action $a$ is selected from a finite subset $A$, and the flow action $b$ is selected from a finite subset $B$. Specifically, for two real numbers satisfying  $ 0 < a_{\min} \leq a_{\max} < 1$, if the queue is non-empty and if the action of the server is $a\in A$, where $A$ is a finite subset of $[a_{\min}, a_{\max}] $, then the service of a customer is successfully completed with probability $a$. Likewise, for two real numbers satisfying  $ 0 \leq b_{\min} \leq b_{\max} < 1$, if the queue is not full and if the action of the server is $b \in B(s)$, where $B(s)$ is a finite subset of $[b_{\min}, b_{\max}]$, then the probability of having one arrival during this time slot is equal to $b$. We assume that $0 \in B(x)$ for all $x$; moreover, when the buffer is full, no arrivals
are possible $(B(L) = {0})$. The transition law $P(\cdot | s,a)$ is therefore given by:
\begin{align*}
        \begin{cases}
       a(1-b) \;\; &\mathrm{if} \;1\leq x\leq L,y = x-1;\\ 
        ab+(1-a)(1-b) \;\; &\mathrm{if} \;1\leq x\leq L,y = x;\\ 
        (1-a)b \;\; &\mathrm{if} \;0 \leq x < L,y = x+1;\\ 
        1-(1-a)b \;\; & \mathrm{if} \; y=x=0;
    \end{cases}
\end{align*}
The reward function $r(s,a,b)$ is a real-valued decreasing function that depends only on $s$, which can be interpreted as a holding cost. The reward function $g^1(s,a,b)$ corresponding to the service rate is assumed to be a decreasing function that depends only on $a$. It can be interpreted as a higher service success rate having a higher cost. The reward function $g^2(s,a,b)$ corresponding to the flow rate $b$ is assumed to be an increasing function that depends only on $b$. It can be interpreted as a higher flow rate is more desired. 

Suppose we want to solve the optimal policy for C-RL problem \eqref{problem:CMDP}, while satisfying constraints for service and flow. In the following numerical example, we compare the result generated by \eqref{eq:SA regularized saddle flow dynamic} and the ground truth result by directly solving the linear programming \ref{eq:LP_CMDP}, where we use the transition law stated above. Specifically, we choose $L = 4, A = [0.2,0.3,0.5,0.6,0.8], B = [0.1,0.3,0.5,0.9,0]$. The initial distribution $q$ is set as uniform distribution. The reward functions are $r(s) = -s+5, g^1(a) = -10a+3, g^2(b) = 10b-3 $. 

\begin{figure}[ht] 
  \label{ fig7} 
  \begin{minipage}[b]{0.5\linewidth}
    \centering
    \includegraphics[width=1\linewidth]{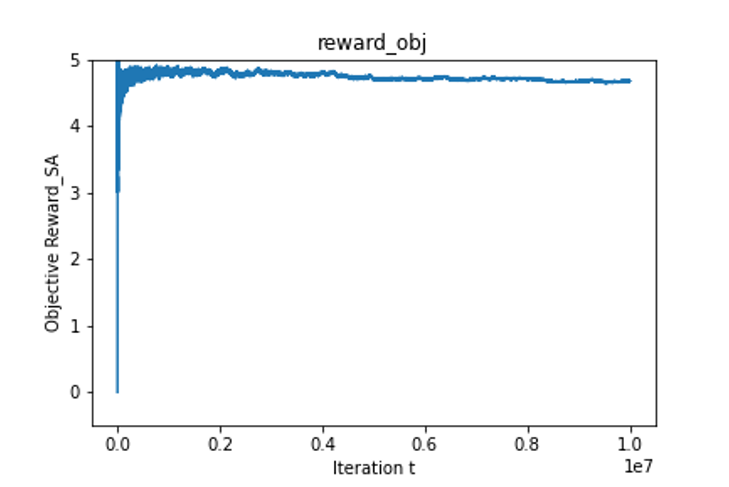} 
    \caption{objective function} 
    \vspace{4ex}
  \end{minipage}
  \begin{minipage}[b]{0.5\linewidth}
    \centering
    \includegraphics[width=1\linewidth]{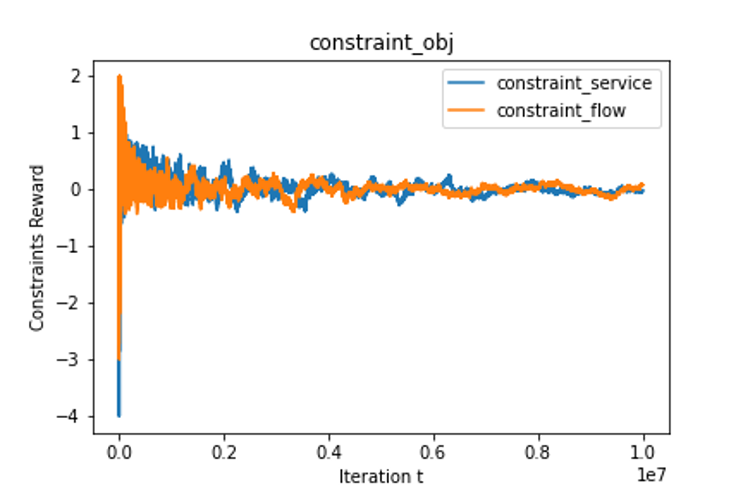} 
    \caption{constraint functions} 
    \vspace{4ex}
  \end{minipage} 
  \begin{minipage}[b]{0.5\linewidth}
    \centering
    \includegraphics[width=1\linewidth]{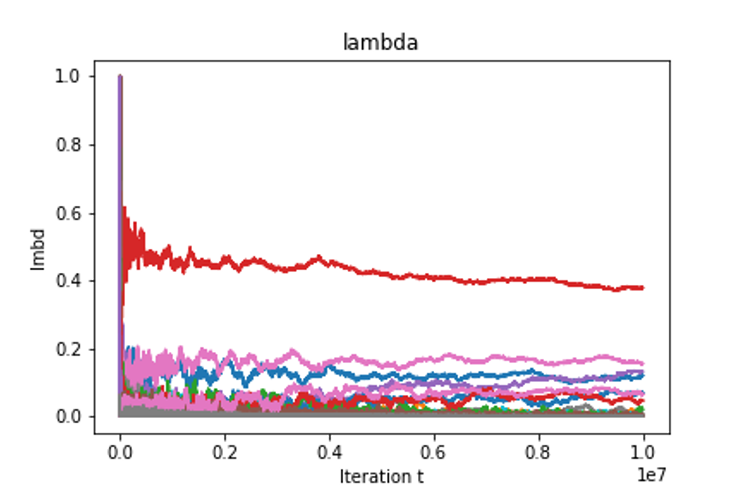} 
    \caption{occupancy measure $\lambda$} 
    \vspace{4ex}
  \end{minipage}
  \begin{minipage}[b]{0.5\linewidth}
    \centering
    \includegraphics[width=1\linewidth]{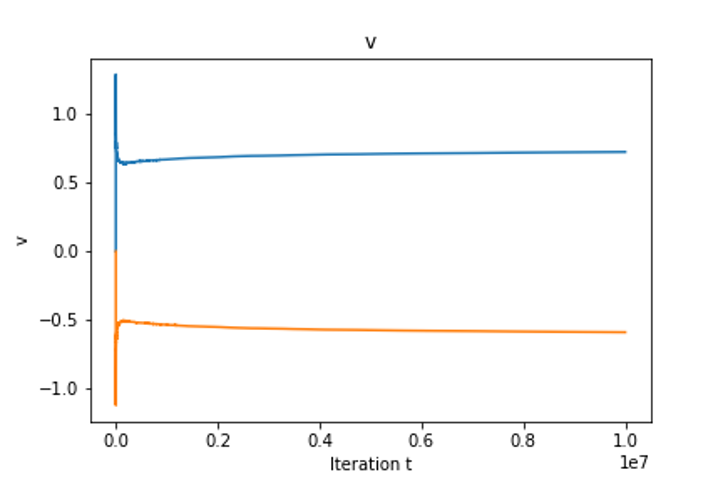} 
    \caption{dual variable v} 
    \vspace{4ex}
  \end{minipage} 
\end{figure}
We compare the cumulative reward function, constraint functions, and output decision variables $\lambda,\mu,v$ with the ground truth result by directly solving the linear programming problem \eqref{eq:LP_CMDP}. Results show that the decision variables converge to the optimal solution while satisfying the constraints for flow and service. 

\section{Conclusion}
In this work, we propose a novel SGDA algorithm to solve the C-RL problem in occupancy measure space leveraging tools from regularized saddle flow dynamics. Even when the Lagrangian function is bilinear, the continuous dynamics asymptotically converge to the optimal occupancy measure and policy. The discretized SGDA is a stochastic approximation of the continuous-time saddle flow dynamic. We further proved the SGDA algorithm almost surely converges to the optimal solution to the C-RL problem.
\ifthenelse{\boolean{arxiv}}{
\appendix
\subsection{Proof of Theorem \ref{thm:Saddle FLow Dynamics}}
We will use the following technical Lemma:
\begin{lem}
For any closed convex set $\mathcal{K} \subset \eucd^n$ and $a, b \in \mathcal{K}, v \in \eucd^n$,the inner product 
\begin{align*}
    \langle b-a,v- \Pi_{ \mathcal{K}}[a,v] \rangle \leq 0
\end{align*}
\textit{Proof:}
According to \cite[Sec.0.6, Cor.1]{aubin2012differential}, we have the following variational inequality holds: 
\begin{align*}
    \langle b- \Psi_{ \mathcal{K}}[c], c - \Psi_{ \mathcal{K}}[c] \rangle \leq 0, \forall b \in \mathcal{K}, \forall c \in \eucd^n.
\end{align*}
The rest follows from \cite[Lem.4]{mallada2008stability}
\end{lem}
Using this lemma, the proof of Theorem \ref{thm:Saddle FLow Dynamics} essentially follows from \cite[Thm.9]{you2021saddle}.
}{}


\bibliographystyle{IEEEtran}

\bibliography{main.bib}

\end{document}